\documentclass[conference]{IEEEtran}
\IEEEoverridecommandlockouts
\usepackage{cite}
\usepackage{amsmath,amssymb,amsfonts}
\usepackage{algorithmic}
\usepackage{graphicx}
\usepackage{booktabs}
\usepackage{multirow}
\usepackage{textcomp}
\usepackage{xcolor}
\newcommand{\modelname}{CLOZE}
\usepackage{algorithm}
\usepackage{algorithmic}
\usepackage{hyperref}
\def\BibTeX{{\rm B\kern-.05em{\sc i\kern-.025em b}\kern-.08em
    T\kern-.1667em\lower.7ex\hbox{E}\kern-.125emX}}
\begin{document}

\title{Utilizing Large Language Models for Zero-Shot \\ Medical Ontology Extension from Clinical Notes}

% \author{\IEEEauthorblockN{Anonymous Authors}
% \IEEEauthorblockA{Paper under double-blind review}}

\author{\IEEEauthorblockN{1\textsuperscript{st} Guanchen Wu}
\IEEEauthorblockA{\textit{Department of Computer Science} \\
\textit{Emory University}\\
guanchen.wu@emory.edu}
\and
\IEEEauthorblockN{2\textsuperscript{nd} Yuzhang Xie}
\IEEEauthorblockA{\textit{Department of Computer Science} \\
\textit{Emory University}\\
yuzhang.xie@emory.edu}
\and
\IEEEauthorblockN{3\textsuperscript{rd} Huanwei Wu}
\IEEEauthorblockA{\textit{College of Public Health} \\
\textit{Temple University}\\
huanmei.wu@temple.edu}
\and
\IEEEauthorblockN{4\textsuperscript{th} Zhe He}
\IEEEauthorblockA{\textit{School of Information} \\
\textit{Florida State University}\\
zhe@fsu.edu}
\and
\IEEEauthorblockN{5\textsuperscript{th} Hui Shao}
\IEEEauthorblockA{\textit{Hubert Department of Global Health} \\
\textit{Emory University}\\
hui.shao@emory.edu}
\and
\IEEEauthorblockN{6\textsuperscript{th} Xiao Hu}
\IEEEauthorblockA{\textit{Nell Hodgson Woodruff School of Nursing} \\
\textit{Emory University}\\
xiao.hu@emory.edu}
\and
\IEEEauthorblockN{7\textsuperscript{th} Carl Yang}
\IEEEauthorblockA{\textit{Department of Computer Science} \\
\textit{Emory University}\\
j.carlyang@emory.edu}
}

\maketitle

% \section*{Abstract}
\begin{abstract}
% Integrating novel medical concepts and relations into existing medical ontologies can substantially improve their coverage and utility in both biomedical research and clinical practice. Traditional rule-based approaches require extensive manual effort, while learning-based methods depend on large annotated datasets and lack adaptability to dynamic hierarchies. Additionally, most current methods lack effective privacy-preserving preprocessing for sensitive clinical texts. However, challenges such as the presence of protected health information (PHI), the complex hierarchical structure of ontologies, and the huge cost of training pose significant limitations to existing methods.To address these issues, we propose a zero-shot ontology extension framework focused on disease ontologies, leveraging Large Language Models (LLMs) for medical entity extraction and relationship inference. The framework combines a dual-agent system—comprising a De-identification Agent and an Entity-extraction Agent—with a hybrid SapBERT–LLM approach to ensure accurate, scalable, and privacy-preserving ontology extension. Experimental results demonstrate the effectiveness of the proposed framework.

% task
Integrating novel medical concepts and relationships into existing ontologies can significantly enhance their coverage and utility for both biomedical research and clinical applications. Clinical notes, as unstructured documents rich with detailed patient observations, offer valuable context-specific insights and represent a promising yet underutilized source for ontology extension.
% challenge
Despite this potential, directly leveraging clinical notes for ontology extension remains largely unexplored.
% our design
To address this gap, we propose \modelname, a novel framework that uses large language models (LLMs) to automatically extract medical entities from clinical notes and integrate them into hierarchical medical ontologies. By capitalizing on the strong language understanding and extensive biomedical knowledge of pre-trained LLMs, \modelname\ effectively identifies disease-related concepts and captures complex hierarchical relationships. The zero-shot framework requires no additional training or labeled data, making it a cost-efficient solution. Furthermore, \modelname\ ensures patient privacy through automated removal of protected health information (PHI).
% experiment and application value
Experimental results demonstrate that \modelname\ provides an accurate, scalable, and privacy-preserving ontology extension framework, with strong potential to support a wide range of downstream applications in biomedical research and clinical informatics.\footnote{The implementation details, prompt designs, and codes are available in the anonymous repository at \href{https://github.com/guanchenwu1015/CLOZE}{https://github.com/guanchenwu1015/CLOZE}.}
\end{abstract}

\section{Introduction}

Ontologies are formal, structured representations of knowledge that define a set of concepts, entities, and the relationships between them within a specific domain~\cite{luschi2023semantic}\cite{xie2025hypkg}. In the biomedical field, for instance, an ontology may represent diseases and symptoms along with their relationships (e.g., \textit{pneumonia} is a type of \textit{respiratory infection}). As medical knowledge evolves, timely ontology extension---systematically identifying, defining, and integrating new entities and their interrelations into the existing structure---is essential for maintaining accurate and up-to-date representations \cite{althubaiti2020combining}. For example, during the early stages of the COVID-19 pandemic, terms such as ``COVID-19'' and related symptoms and complications had to be rapidly incorporated into ontologies like SNOMED CT to support reliable clinical interpretation. Without such updates, downstream systems risk relying on incomplete knowledge, leading to degraded performance and potential clinical harm. Scalable and accurate ontology extension is therefore critical for ensuring the effectiveness of biomedical applications.

Clinical notes, which are unstructured textual documents produced during patient encounters—including discharge summaries, nursing records, radiology reports, and ECG interpretations—represent a rich source of medical knowledge \cite{seinen2025using}. These narratives capture detailed clinical observations, treatment rationales, patient interactions, and social determinants of health—information frequently absent or poorly represented in other clinical documents such as structured Electronic Health Records (EHRs) \cite{tayefi2021challenges}\cite{xie2025kerap}. As a result, clinical notes offer valuable, context-specific insights that are highly relevant for ontology extension. By integrating novel medical concepts derived from these narratives, ontologies can achieve greater coverage, granularity, and practical applicability. 
Despite their potential, directly leveraging clinical notes for ontology extension remains largely unexplored, resulting in the loss of valuable clinical information embedded within these unstructured texts.

% Existing ontology extension methods can be broadly classified into two categories: (i) rule-based and statistical methods, which extract new entities and relationships using patterns, co-occurrence analysis, and heuristic rules \cite{cruanes2011ontology, behr2023ontology}; and (ii) learning-based methods, which leverage machine learning methods to predict and integrate new concepts from structured or unstructured data sources \cite{song2014novel, huang2019hierarchical, pesquita2012predicting}. 

To leverage clinical notes for ontology extension, two key steps are essential: (i) Medical Entity Extraction, which identifies medical entities from unstructured clinical text, and (ii) Ontology Extension, which determines the appropriate insertion point for each extracted entity within an existing ontology. Existing approaches to ontology extension typically fall into two categories:
(i) Rule-based and statistical methods, which rely on predefined patterns, co-occurrence analysis, and heuristic rules to extract new entities and relationships~\cite{cruanes2011ontology, behr2023ontology}; and
(ii) Learning-based methods, which apply machine learning techniques to infer and integrate new concepts from structured or unstructured data. However, these methods face significant limitations when applied to clinical notes. Rule-based and statistical approaches require substantial manual effort and lack scalability for large, evolving ontologies. They also perform poorly in capturing complex hierarchies—for instance, correctly placing ``triple-negative breast cancer'' demands nuanced domain understanding beyond simple lexical patterns. Learning-based methods, on the other hand, depend heavily on large annotated datasets and often lack interpretability, making it difficult to validate updates. Moreover, applying such models to clinical texts raises privacy concerns, as sensitive patient information may be inadvertently exposed.

\begin{figure}[t]
\centering
\includegraphics[width=\linewidth]{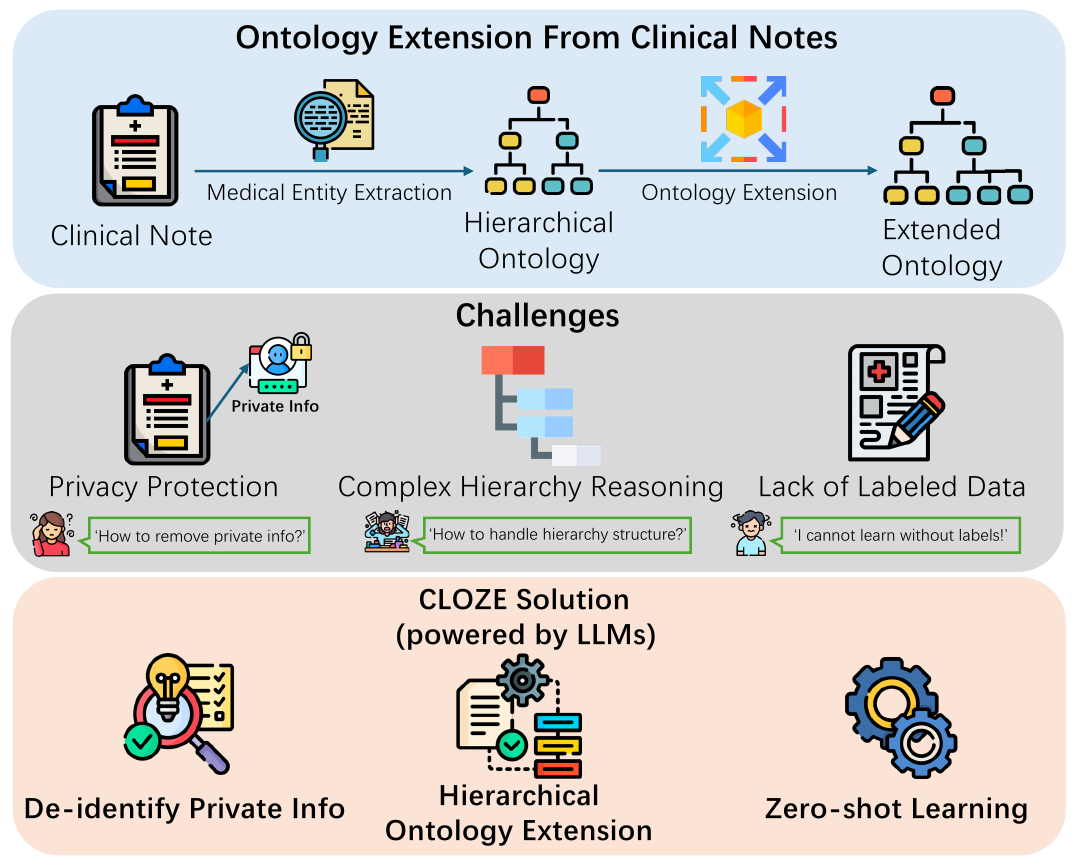}
\caption{Illustration of existing ontology extension methods and our proposed framework.}
\label{fig:1}
\vspace{-7mm}
\end{figure}

In this work, we propose \modelname\ (\underline{CL}inical Notes \underline{O}ntology \underline{Z}ero-shot \underline{E}xtension), a novel framework for hierarchical ontology extension from clinical notes, as illustrated in Figure~\ref{fig:1}. To the best of our knowledge, this is the first framework designed specifically for hierarchical ontology extension using clinical notes. The design of \modelname\ is motivated by three core needs: 
(i) applying robust de-identification methods to protect patient privacy throughout the pipeline, 
(ii) leveraging the reasoning capabilities of LLMs to accurately capture complex hierarchical relationships that traditional approaches often overlook, and 
(iii) harnessing the rich biomedical knowledge embedded in pretrained language models to support a fully end-to-end, zero-shot pipeline that eliminates the need for manual annotation or training, thereby overcoming data scarcity in biomedical domains.

\modelname\ consists of two main steps: (i) Medical Entity Extraction and (ii) Hierarchical Ontology Extension. In the Medical Entity Extraction step, \modelname\ first uses an LLM as the De-identification Agent to remove PHI (Protected Health Information), aligning with standard privacy regulations (e.g., HIPAA) to ensure safe handling of private data. Then another LLM acts as the Entity-extraction Agent, identifying candidate medical entities from de-identified notes. In the Hierarchical Ontology Extension step, these entities are further integrated into an existing ontology. Specifically, we employ a pretrained biomedical language model to compute semantic embeddings of new entities and existing ontology nodes, enabling the identification of the most semantically relevant anchor points in the ontology. Another LLM serves as the Relation-determination Agent, inferring the relationships between new and existing concepts. Based on this classification, \modelname\ recursively inserts new entities at the appropriate hierarchical level, preserving both structural coherence and semantic fidelity.

In our experiments, we evaluate \modelname\ using clinical notes from a major U.S. hospital, focusing on disease-related entities for ontology extension. We compare performance against state-of-the-art (SOTA) baseline models to validate \modelname. Results show that \modelname\ effectively and efficiently enables privacy-preserving processing and scalable, zero-shot ontology extension with high accuracy in complex biomedical settings.

\section{Related Work}
% Recent advancements in ontology extension methodologies can be broadly categorized into two groups: (i) rule-based and statistical methods, and (ii) learning-based methods. 

\subsection{Rule-based and Statistical Methods for Ontology Extension}
Traditional ontology extension techniques often rely on rule-based systems and statistical analysis to extract new entities, relationships, and hierarchical structures from domain-specific corpora. For example, in the medical treatment domain,  Cruanes et al. applied semantic tagging and text mining to extract and integrate specialized terms, enabling structured updates to medical ontologies \cite{cruanes2011ontology}.
% In the catalytic sciences domain, Behr et al. used information extraction tools to identify new categories and automate ontology expansion \cite{behr2023ontology}. 
Phrase2Onto demonstrates the use of phrase-based topic modeling to detect novel concepts in free text and integrate them into ontologies, highlighting the effectiveness of unsupervised statistical techniques in real-world applications \cite{pour2023phrase2onto}. Co-occurrence analysis and word filtering have also been applied to identify candidate entities, followed by subsumption analysis to embed them within shallow ontology hierarchies \cite{he2013hybrid}.

% Ontology extension \cite{santosa2021flat}. Flat classification treats all categories independently, achieving higher performance in simple tasks but lacking scalability and fine-grained control \cite{althubaiti2020combining}. In contrast, hierarchical classification respects ontological structure and taxonomy levels, though its effectiveness can be hindered by label imbalance and increased target complexity \cite{huang2019hierarchical}.

While rule-based and statistical methods are often interpretable and lightweight, they typically require significant manual tuning, domain-specific knowledge, and validation by experts. These factors limit their scalability and adaptability.

\subsection{Learning-based Methods for Ontology Extension}
Learning-based approaches, including traditional machine learning and deep learning models, offer data-driven alternatives to rule-based systems for ontology extension. Early efforts used supervised classifiers (such as Bayesian models) and clustering techniques  (such as k-means) to discover and integrate new concepts into domain ontologies \cite{song2014novel}. Hybrid pipelines combining search engines, multilayer perceptrons (MLPs), Naive Bayes, and neural networks were also proposed to infer relationships between candidate entities \cite{he2013hybrid}.

With the rise of deep learning, more advanced models have been introduced to capture semantic and structural relationships. Word embeddings, such as Word2Vec, combined with hierarchical clustering, have enabled the automatic grouping of similar biomedical terms \cite{behr2023ontology}. Attention-based models and visual explanation mechanisms have also been employed to improve interpretability and concept classification in chemistry and biomedicine \cite{memariani2021automated}. In the biomedical domain, Pesquita and Couto proposed a supervised method that predicts which regions of an ontology are likely to require extension by analyzing historical versions and external annotations \cite{pesquita2012predicting}.

Despite their promise, learning-based methods often depend on large, annotated datasets and tend to be difficult to interpret, which complicates the validation of their outputs. Furthermore, when applied to clinical text, these models may inadvertently compromise patient privacy by exposing sensitive information.

% \subsection{Entity Recognition and Extraction}
% Traditional Named Entity Recognition (NER) methods—commonly used as preprocessing for ontology extension—face limitations in clinical applications. These models, often trained on general-purpose corpora or narrow biomedical datasets, struggle to generalize to the complex and variable language of clinical narratives. The challenge is further amplified in de-identified texts, where key contextual cues are removed. Moreover, conventional NER systems are typically designed to identify a fixed set of entity types, making them poorly suited for discovering novel or fine-grained entities necessary for ontology expansion.

\subsection{Pre-trained Language Models (PLMs)}
Considering the prohibitive cost of training from scratch, researchers have turned to pretrained language models (PLMs)—such as BERT and GPT—which leverage massive corpora to produce rich, context‐sensitive embeddings. In the biomedical domain, further specialization has led to models such as BioBERT \cite{lee2020biobert} and SapBERT \cite{liu2020self}. SapBERT, in particular, incorporates a self-alignment objective to preserve semantic similarity within its embeddings, making it one of the SOTA models for generating semantic representations used in biomedical ontology alignment.

More recently, LLMs have demonstrated strong zero‐ and few‐shot performance on entity extraction and ontology extension tasks \cite{xie2024promptlink}\cite{bhasuran2025preliminary}. For example, Wu et al. \cite{wu2024ontology} proposed a zero‐shot extension framework that combines LLM agents with online hierarchical clustering to integrate medical symptom concepts from patient forums. However, LLM‐based approaches often struggle with deeper, multi‐layer ontologies \cite{tsaneva2024llm}. Their flat textual inputs and limited context windows impede the explicit modeling of hierarchical structure, leading to hallucinations or misclassifications when scaling beyond a few ontology levels.

% In summary, learning-based methods have significantly improved the scalability and automation of ontology extension workflows. However, standalone NER or LLM-based approaches remain limited when applied to complex biomedical ontologies—especially in the presence of de-identified or deeply structured input. 
\section{Method}

% \subsection{Problem Formulation}
% In this study, we address the task of extracting novel medical entities from clinical notes and integrating them into existing hierarchical medical ontologies. An ontology \(O=(E,R)\) is a depth-\(k\) hierarchy of entities \(E\) and their interrelationships \(R\), providing a shared conceptual model of a domain. Our goal is to extend \(O\) with newly discovered entities from clinical notes—without any additional task-specific training data—thereby improving the scalability and applicability of ontology extension. The problem is formally defined as follows:

% \begin{itemize}
%     \item Input: A collection of clinical notes denoted as \(D = \{D_1, D_2, \dots, D_N\}\), where \(D_i\) represents the \(i\)-th clinical note and \(N\) is the total number of notes. In addition, we are given a hierarchical medical ontology \(O = (E, R)\), where \(E\) is the set of entities and \(R\) is the set of relationships among them. The ontology has a hierarchical depth of \(k\).
    
%     \item Output: An extended medical ontology \(\Tilde{O} = (\Tilde{E}, \Tilde{R})\), where \(\Tilde{E}\) and \(\Tilde{R}\) denote the updated sets of entities and relationships, respectively. The goal is to incorporate newly extracted entities from \(D\) into the hierarchical structure of \(O\), while preserving ontological consistency.
% \end{itemize}

\subsection{Problem Formulation}

We address the task of extracting novel medical entities from clinical notes and integrating them into an existing hierarchical ontology in a privacy-preserving and zero-shot manner. Let \( D = \{D_1, D_2, \dots, D_N\} \) denote a collection of clinical notes, where \( D_i \) is the \(i\)-th document and \(N\) is the total number of notes.

We are also given a medical ontology \( O = (E, R, \prec) \), where \( E \) is the set of existing medical entities, \( R \subseteq E \times E \) is the set of ontological relationships (e.g., ``is-a'', ``part-of''), and \( \prec \) denotes a partial order capturing the hierarchical structure with depth \( k \). 

The goal is to extract a set of novel medical entities \( E^{\text{new}} = \{e_1, e_2, \dots, e_M\} \) from \(D\), where \(e_j \notin E\), and to generate candidate mappings \( A = \{(e_j, e_p) \mid e_p \in E\} \) such that each new entity \( e_j \) is assigned a parent \( e_p \) in the ontology hierarchy. The final output is an extended ontology \( \Tilde{O} = (\Tilde{E}, \Tilde{R}, \prec) \), where:

\begin{itemize}
    \item \( \Tilde{E} = E \cup E^{\text{new}} \) is the updated set of entities.
    \item \( \Tilde{R} = R \cup R^{\text{new}} \), where \( R^{\text{new}} \subseteq E^{\text{new}} \times E \) represents the set of newly established relationships based on predicted parent-child mappings.
    \item \( \prec \) is preserved or updated to maintain a valid hierarchy of depth \( \Tilde{k} \geq k \).
\end{itemize}

The entire process is conducted without access to any task-specific labeled data or manually annotated ontological links, thereby enabling scalable and privacy-preserving ontology extension in a zero-shot setting.

% \textit{\textbf{\underline{Framework Overview.}}}
\subsection{Framework Overview}

\begin{figure*}[t!]
\centering
\includegraphics[scale=0.6]{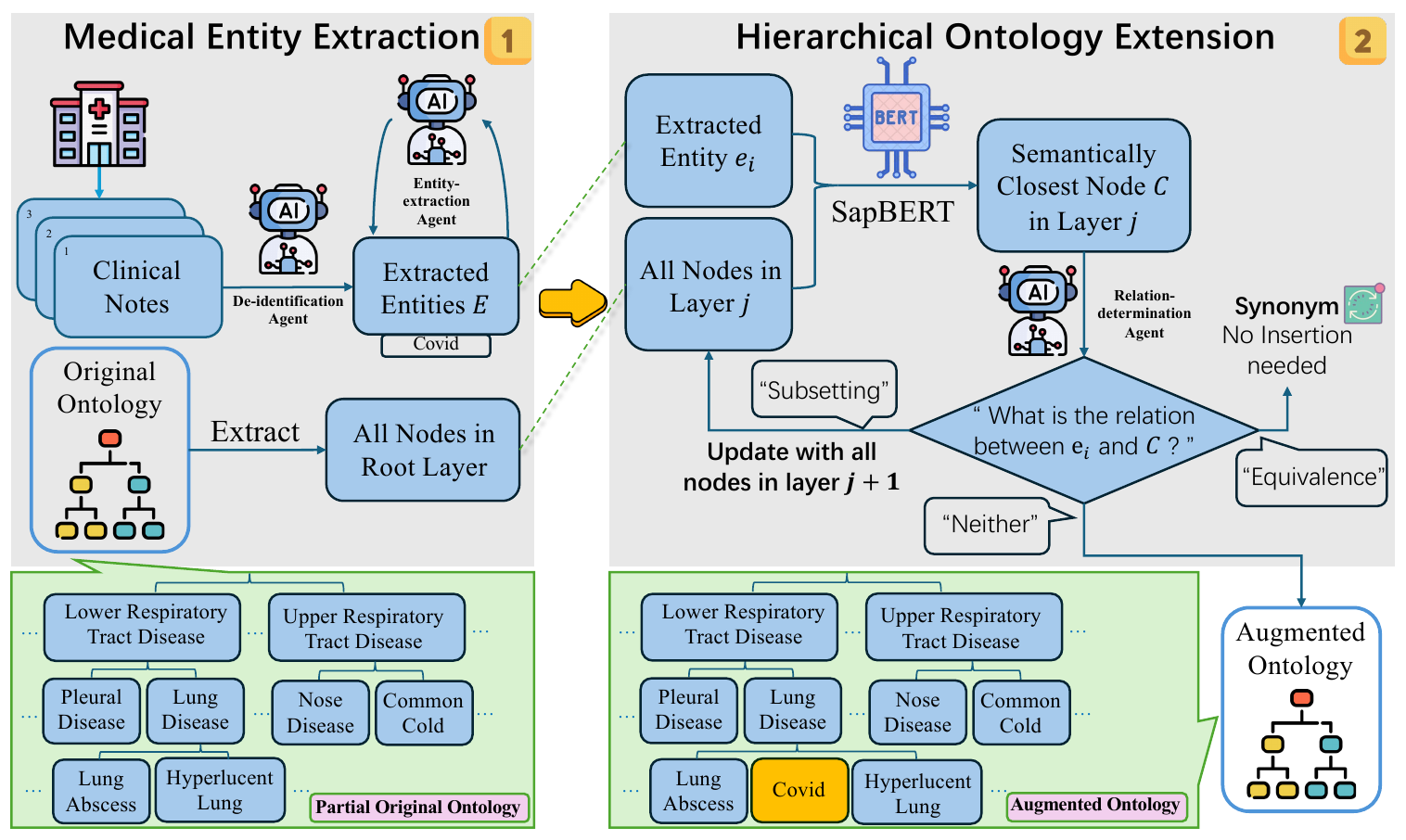}

% \caption{Framework Overview. This framework has two components: Private Data Preprocessing and Hierarchical Ontology Extension. The first module de-identifies PHI in clinical notes and extracts medical entities using two LLM agents. The second module integrates these entities into the ontology, using SapBERT for semantic matching and an LLM to determine relationships. The process iterates layer by layer until integration is complete.}
\caption{Framework of CLOZE. This framework has two components: Medical Entity Extraction and Hierarchical Ontology Extension. The first module de-identifies PHI in clinical notes and extracts medical entities using two LLM agents. The second module integrates these entities into the ontology, using SapBERT for semantic matching and an LLM to determine relationships. The process iterates layer by layer until integration is complete.}
\label{fig:2}
\vspace{-4mm}
\end{figure*}

To support the integration of novel medical entities into hierarchical ontologies, we propose \modelname, a two-step framework: (i) Medical Entity Extraction and (ii) Hierarchical Ontology Extension, as illustrated in Figure \ref{fig:2}. The extraction step uses two LLM-powered agents: a \textit{De-identification Agent} to remove PHI and ensure privacy, and an \textit{Entity-extraction Agent} to identify medical terms from the de-identified text. In the extension step, a \textit{PLM (SapBERT)} generates embeddings for new entities and candidate ontology nodes, while a \textit{Relation-determination Agent} classifies each pairwise relation to recursively insert new entities at the appropriate hierarchical level, preserving structural and semantic consistency. Further details are provided in Sections \ref{sec:step1} and \ref{sec:step2}.

\subsection{Medical Entity Extraction}
\label{sec:step1}

The Medical Entity Extraction step in \modelname\ consists of two LLM-powered agents designed to ensure both privacy preservation and comprehensive entity identification. Let \( D = \{D_1, D_2, \dots, D_N\} \) denote the set of raw clinical notes, each potentially containing protected health information (PHI).

\paragraph{De-identification Agent}
To comply with privacy regulations, we first apply a De-identification Agent to each note \( D_i \), producing a de-identified version \( D_i^{\text{deid}} \). Following recent advances in LLM-based PHI de-identification \cite{wu2025towards, wu2025large}, this step removes personal identifiers before downstream processing and is defined as:
\[
D_i^{\text{deid}} = f_{\text{deid}}(D_i),
\]
where \( f_{\text{deid}} \) is an LLM prompted to extract a predefined set of PHI types: 
\begin{multline*}
\mathcal{P} = \{\,\texttt{person},\ \texttt{location},\ \texttt{organization},\ \texttt{age},\\
\texttt{phone number},\ \texttt{email},\ \texttt{date and time},\\
\texttt{zip},\ \texttt{profession},\ \texttt{username},\ \texttt{id},\ \texttt{url}\,\}
\end{multline*}

from the clinical notes. The LLM agent outputs a JSON dictionary:
\[
\texttt{PHI\_dict} = \{ p : [e_1^p, e_2^p, \dots] \mid p \in \mathcal{P} \},
\]
where each \( e_j^p \) is an extracted entity of PHI type \( p \). This structured output improves interpretability, enables traceability, and promotes high recall, including for borderline PHI instances. Finally, the de-identified note $D_i^{\text{deid}} $ is constructed by ``masking'' these detected PHIs in the raw clinical notes. 

\paragraph{Entity-extraction Agent}

After de-identification, the Entity-extraction Agent processes the de-identified clinical note \( D_i^{\text{deid}} \) to extract disease-related medical concepts. This step is defined as:
\[
E^{\text{new}} = f_{\text{extract}}(D_i^{\text{deid}}),
\]
where \( E^{\text{new}} = \{e_1^{\text{new}}, e_2^{\text{new}}, \dots, e_m^{\text{new}}\} \) denotes the set of disease entities extracted from the note. The extraction function \( f_{\text{extract}} \) is implemented as a large language model (LLM) guided by a structured prompt that emulates the behavior of a clinical expert. The prompt is carefully designed to: (i) frame the task specifically as disease-only entity recognition, (ii) instruct the model to return a flat, deduplicated list of disease mentions. This step provides the foundation for downstream ontology integration by producing a curated list of candidate entities for insertion into the ontology.

\subsection{Hierarchical Ontology Extension}
\label{sec:step2}

After extracting de-identified medical entities, the next step in our framework is to integrate them into an existing hierarchical ontology while preserving semantic meaning and structural consistency. 
We define the existing ontology as \( O = (E, R, \prec) \), where \( E \) is the set of ontology nodes (entities), \( R \subseteq E \times E \) represents the hierarchical relationships (e.g., ``is-a''), and \( \prec \) defines the partial ordering over nodes that encodes the ontology’s hierarchical structure.

Let \( e^{\text{new}} \in E^{\text{new}} \) denote a newly extracted entity from the clinical notes. To determine its appropriate insertion point within the ontology, we adopt a hybrid strategy that combines (i) PLM-based semantic matching, (ii) a Relation-determination Agent for relational classification, and (iii) recursive hierarchical extension. The goal is to identify an anchor node \( e^* \in E \) that is most semantically similar to \( e^{\text{new}} \), and to assess whether the new entity is either semantically equivalent to, or a more specific subclass of, \( e^* \) or one of its descendants. If no suitable child node is identified through recursive extension, which means \( e^{\text{new}} \) is a new found ``child'' we extracted from the clinical notes, then \( e^{\text{new}} \) is inserted as a new ``child'' node under candidate \( e^* \), thereby maintaining both semantic alignment and structural consistency in the extended ontology.

\paragraph{PLM-based Matching}
We leverage SapBERT, a SOTA PLM for generating semantic representations in biomedical ontology alignment, to compute embeddings for both the new entity and candidate ontology nodes. Specifically, we obtain an embedding \( \mathbf{v}_{e^{\text{new}}} \in \mathbb{R}^d \) for the name of the new entity, and an embedding \( \mathbf{v}_{e} \in \mathbb{R}^d \) for the name of each candidate node \( e \in E_l \), where \( E_l \subseteq E \) denotes the set of nodes in the current ontology layer \( l \). Cosine similarity is then used to identify the most semantically similar candidate:
\[
e^* = \arg\max_{e \in E} \cos\left(\mathbf{v}_{e^{\text{new}}}, \mathbf{v}_{e}\right).
\]

\paragraph{Relation-determination Agent}
Given the candidate node \( e^* \), the Relation-determination Agent—an LLM prompted with contextual descriptions of \( e^{\text{new}} \) and \( e^* \)—classifies their relationship as one of the following:
\begin{itemize}
  \item \texttt{Equivalence}: \( e^{\text{new}} \equiv e^* \), indicating they refer to the same medical concept. The entity is treated as a synonym and is not added.
  \item \texttt{Subsetting}: \( e^{\text{new}} \prec e^* \), meaning the new entity is more specific and should be inserted as a child of \( e^* \).
  \item \texttt{Neither}: The entities are unrelated in a hierarchical sense, prompting the search to continue elsewhere in the ontology.
\end{itemize}

\paragraph{Recursive Hierarchical Extension}
If the relationship between the new entity and the matched candidate node \( e^* \) is classified as \texttt{subsetting}, the process recurses into the next layer of the ontology, using the children of \( e^* \) as the new candidate set \( E_{l+1} \). This recursive search continues layer by layer until one of the following occurs:

\begin{itemize}
    \item The relation is classified as \texttt{equivalence}, indicating that \( e^{\text{new}} \) already exists in the ontology and no insertion is needed.
    \item The relation is classified as \texttt{subsetting}, indicating that \( e^{\text{new}} \) should be put to the next layer, the search moves to another layer recursively.
    \item No child nodes of \( e^* \) are classified as \texttt{subsetting} or \texttt{equivalence}, suggesting that \( e^{\text{new}} \) is more specific than \( e^* \) but not semantically related to any of its children. In this case, \( e^{\text{new}} \) is treated as a new ``child'' node under \( e^* \).
\end{itemize}

In either case where insertion is required, the new entity is added as a child of the most recent parent node:
\[
\Tilde{E} = E \cup \{e^{\text{new}}\}, \quad
\Tilde{R} = R \cup \{(e^{\text{new}}, e_{\text{parent}})\}.
\]
The algorithm of the Hierarchical Ontology Extension step is summarized in Algorithm \ref{alg:recursive-insert}. By combining SapBERT’s high-quality biomedical embeddings with the contextual reasoning abilities of the LLM, our framework enables accurate and scalable ontology extension. This hybrid approach ensures that new entities are integrated into deep hierarchical structures with both semantic alignment and structural consistency.

\begin{algorithm}[htbp]
\caption{Hierarchical Ontology Extension}
\label{alg:recursive-insert}
\begin{algorithmic}[1]
\REQUIRE New extracted entity \( e^{\text{new}} \), top-layer candidate set \( E_0 \)
\ENSURE Updated ontology \( \Tilde{O} = (\Tilde{E}, \Tilde{R}) \)

\STATE Set layer index \( l \leftarrow 0 \)
\WHILE{true}
    \STATE Compute embedding \( \mathbf{v}_{e^{\text{new}}} \) and embeddings \( \mathbf{v}_{e} \) for all \( e \in E_l \)
    \STATE Identify closest match:
    \[
    e^* \leftarrow \arg\max_{e \in E_l} \cos(\mathbf{v}_{e^{\text{new}}}, \mathbf{v}_{e})
    \]
    \STATE Query Relation-determination Agent for relation between \( e^{\text{new}} \) and \( e^* \)
    \IF{relation is \texttt{equivalence}}
        \STATE \textbf{return} No insertion needed
    \ELSIF{relation is \texttt{subsetting}}
        \STATE Set \( E_{l+1} \leftarrow \text{Children}(e^*) \)
        \STATE Increment layer: \( l \leftarrow l + 1 \)
    \ELSE
        \STATE Insert \( e^{\text{new}} \) as child of \( e^* \)
        \STATE Update:
        \[
        \Tilde{E} \leftarrow E \cup \{e^{\text{new}}\}, \quad
        \Tilde{R} \leftarrow R \cup \{(e^{\text{new}}, e^*)\}
        \]
        \STATE \textbf{return} Updated ontology \( \Tilde{O} = (\Tilde{E}, \Tilde{R}) \)
    \ENDIF
\ENDWHILE
\end{algorithmic}
\end{algorithm}

\section{Experimental Settings}

\subsection{Dataset}
Access to real-world clinical notes is highly restricted due to stringent privacy regulations, and fully annotated clinical datasets are particularly costly and difficult to obtain. For this study, we obtained a dataset consisting of 100 fully annotated clinical notes, in which all PHI entities were meticulously identified by multiple medical experts. These notes were generously provided by a large hospital in the U.S., ensuring real-world clinical authenticity and high annotation quality. These notes retain their original structure and include a diverse and detailed range of patient information, making them a rare and invaluable resource for advancing research in clinical informatics. On average, each clinical note contains approximately 1,000 tokens, reflecting a substantial level of detail and comprehensiveness in documenting patient care. 
% The clinical notes encompass multiple aspects, including medical history, clinical findings, diagnostic evaluations, and treatment plans. This level of detail ensures that this dataset captures a broad spectrum of clinical scenarios, making it well-suited for evaluating our proposed framework. The dataset encompasses various categories of medical entities, including diseases, symptoms, and medications. A detailed analysis of the dataset revealed that disease entities are more prevalent compared to other categories. 
Our goal is to extract the disease entities from the notes and integrate them into the hierarchical structure of a disease-specific ontology.

As a foundational resource, the Disease Ontology (DO) was utilized as the seed ontology \cite{baron2024kb}. DO is a standardized biomedical ontology offering a comprehensive and hierarchical classification of human diseases. It integrates seamlessly with major medical vocabularies and databases, including ICD and UMLS, thereby facilitating robust alignment and analysis of disease-related data. Widely regarded as a critical tool in biomedical research and healthcare, DO supports data integration, analysis, and interoperability across diverse systems, making it an essential resource for advancing ontology-based methodologies in the medical domain.

\subsection{Language Models}
For the \modelname\ pipeline, we utilize one pre-trained language model (SapBERT) and three LLM-based agents (De-identification Agent, the Entity-extraction Agent, and the Relation-determination Agent). The SapBERT model is obtained from the Hugging Face platform\footnote{\href{https://huggingface.co/cambridgeltl/SapBERT-UMLS-2020AB-all-lang-from-XLMR}{https://huggingface.co/cambridgeltl/SapBERT-UMLS-2020AB-all-lang-from-XLMR}} and is used to generate biomedical entity embeddings for semantic matching. All three LLM-based agents are built upon the LLaMA-3-70B-Instruct model\footnote{\href{https://huggingface.co/meta-llama/Meta-Llama-3-70B-Instruct}{https://huggingface.co/meta-llama/Meta-Llama-3-70B-Instruct}}, a SOTA model from Meta’s LLaMA series. These language models are deployed locally, ensuring that sensitive clinical data remains entirely on-premises. This setup effectively addresses privacy concerns and supports compliance with data protection standards. The combination of local deployment and advanced language understanding enables efficient and confidential processing of clinical notes.

For the evaluation in Section \ref{sec:result2}, considering the lack of ground truth, we also incorporated the GPT-4-0613 model\footnote{\href{https://platform.openai.com/docs/models/gpt-4}{https://platform.openai.com/docs/models/gpt-4}} via Microsoft’s Azure OpenAI Service\footnote{\href{https://azure.microsoft.com/en-us/products/ai-services/openai-service}{https://azure.microsoft.com/en-us/products/ai-services/openai-service}}. Azure is a cloud-based platform provides robust safeguards, including end-to-end encryption, isolated execution environments, and HIPAA-compliant infrastructure. These features ensure that both prompt inputs and model outputs remain secure and inaccessible to external parties, including OpenAI. By leveraging Azure’s secure environment, we were able to utilize the powerful reasoning capabilities of GPT models while maintaining the confidentiality of PHI. This configuration offers a reliable and compliant solution for deploying language model inference in real-world clinical contexts.

\subsection{Baseline Methods}
\label{sec:baselines}
Since no existing work directly addresses ontology extension from clinical notes, we evaluate our framework by assessing each component separately. 

In Section~\ref{sec:result1}, we evaluate the two agents (De-identification Agent and Entity-extraction Agent) in the Medical Entity Extraction step, each against relevant baseline methods. For the De-identification Agent, we exclude traditional learning-based approaches due to the limited availability of large annotated datasets required for training. Instead, we compare our design using LLaMA-3-70B-Instruct against the following zero-shot baselines for PHI annotation:
\begin{itemize}
    \item PhysioNet: A widely used rule-based method, implemented via the PhysioNet de-identification software package\footnote{\href{https://www.physionet.org/content/deid/1.1/}{https://www.physionet.org/content/deid/1.1/}}, which uses regular expressions and dictionary-based lookups to identify and redact PHI in clinical text~\cite{neamatullah2008automated}.
    \item LLaMA-3-8B-Instruct: A compact 8B-parameter version\footnote{\href{https://huggingface.co/meta-llama/Meta-Llama-3-8B-Instruct}{https://huggingface.co/meta-llama/Meta-Llama-3-8B-Instruct}} of Meta’s LLaMA-3 series, fine-tuned for instruction-following tasks and efficient inference with limited computational resources.    
    \item GPT-3.5-turbo-0301: A cost-effective version\footnote{\href{https://platform.openai.com/docs/models/gpt-3.5-turbo}{https://platform.openai.com/docs/models/gpt-3.5-turbo}} of the GPT-3.5 family, optimized for fast responses and general-purpose natural language understanding tasks, including few-shot and zero-shot reasoning.
    \item GPT-4-0613: A state-of-the-art LLM known for its superior reasoning, comprehension, and instruction-following capabilities. The 0613 version\footnote{\href{https://platform.openai.com/docs/models/gpt-4}{https://platform.openai.com/docs/models/gpt-4}} is tailored for stable function-calling and prompt engineering in NLP tasks.
\end{itemize}

For the Entity-extraction Agent, we identify disease-related entities from de-identified clinical notes. For both agents, we evaluate the extracted disease entities using a standardized vocabulary from the Disease Ontology, ensuring consistent and ontology-aligned comparisons across methods. We compare our LLM-based approach (LLaMa-3-70B-Instruct) against:
  \begin{itemize}
    \item Stanza~\cite{qi2020stanza}: A widely adopted biomedical NLP toolkit\footnote{\href{https://stanfordnlp.github.io/stanza/}{https://stanfordnlp.github.io/stanza/}} that performs named entity recognition on clinical text.
    \item BioEN~\cite{raza2022large}: A recent model\footnote{\href{https://huggingface.co/MilosKosRad/BioNER}{https://huggingface.co/MilosKosRad/BioNER}} based on fine-tuned DistilBERT, trained on the MACCROBAT 2020 dataset. It is designed to extract a wide range of biomedical and epidemiological entities without relying on external ontologies for concept normalization.
  \end{itemize}

In Sections~\ref{sec:result2} and~\ref{sec:result3}, we evaluate the performance of our Hierarchical Ontology Extension step. We compare our full method (SapBERT + LLM-Hierarchical) against the following three baseline strategies:
\begin{itemize}
    \item LLM-Onetime: This method prompts a single large language model to perform ontology extension in one step. Due to context window limitations, the ontology is divided into segments. For each segment, the LLM identifies the node most similar to the new entity. The node with the highest overall similarity is selected, and its parent-child structure is used to determine the insertion point.
    
    \item LLM-Hierarchical: This baseline simulates hierarchical extension by prompting the LLM with one ontology layer at a time. At each step, the model selects the most semantically similar node and classifies the relationship as \texttt{subsetting}, \texttt{equivalence}, or \texttt{neither}. If classified as \texttt{subsetting}, the search continues recursively into the children of the selected node.
    
    \item SapBERT + LLM-Onetime: This variant combines SapBERT and a single-step LLM prompt. SapBERT is used to compute embeddings for both the new entity and all ontology nodes. The node with the highest cosine similarity is selected, and an LLM is prompted once to classify the relationship between the new entity and the selected node.
\end{itemize}

\subsection{Evaluation Metrics}
\label{sec:metrics}
In Section~\ref{sec:result1}, we report precision, recall, and F1 score to evaluate the effectiveness of our design in the Medical Entity Extraction step. For the experiment evaluating the De-identification Agent, the ground truth consists of manually annotated PHI entities in the dataset. For the Entity-extraction Agent, an approximate ground truth is constructed by scanning each clinical note and identifying Disease Ontology (DO) terms that appear in the text using fuzzy string matching with a high similarity threshold (e.g., 90). These matched DO terms serve as the reference set. Predicted entities from each method are then compared against this reference using a range of relaxed fuzzy thresholds (e.g., from 60 to 80) to assess robustness under varying levels of string similarity.

% Precision is calculated as the number of matched predicted entities divided by the total number of predicted entities. Recall is the number of matched predicted entities divided by the total number of ground truth entities. F1 score is the harmonic mean of precision and recall. 

In Section~\ref{sec:result2}, considering the lack of ground truth, we evaluate the Hierarchical Ontology Extension step using GPT-4-0613 for automatic assessment against baseline methods. For each updated ontology, the language model classifies every newly extended entity relationship as ``Correct,'' ``Incorrect,'' or ``Not Sure.'' Precision is computed as the number of correct predictions divided by the total number of labeled predictions, excluding those marked as uncertain. This precision metric serves as the primary indicator of how accurately each method integrates new entities into the ontology structure.

In Section~\ref{sec:result3}, we further evaluate the effectiveness of the Hierarchical Ontology Extension step through human assessment. A random sample of newly inserted entity relationships is manually reviewed based on four criteria: relevance (whether the entity is semantically appropriate within the disease ontology), accuracy (whether the hierarchical relationship is correctly specified), importance (whether the new entity adds meaningful value to the ontology), and overall satisfaction (the overall quality and consistency of the integrated triplet).
\section{Experimental Results}

\subsection{Medical Entity Extraction}
\label{sec:result1}
We begin by evaluating the performance of our two agents (the De-identification Agent and the Entity-extraction Agent) in the Medical Entity Extraction step. Results are presented in Table~\ref{tab:1} and Table~\ref{tab:2}, and compared against the baseline methods introduced in Section~\ref{sec:baselines} using the evaluation metrics defined in Section~\ref{sec:metrics}.

\vspace{-1em}
\begin{table}[htbp]
\centering
\small
\renewcommand{\arraystretch}{1.1}
\caption{\textbf{De-identification Agent evaluation results.} Precision, Recall, and F1 score across different models. Best results are bolded.}
\label{tab:1}
\begin{tabular}{cccc}
\toprule
\textbf{Method} & \textbf{Precision} & \textbf{Recall} & \textbf{F1} \\
\midrule
PhysioNet            & 0.39 & 0.24 & 0.28 \\
GPT-3.5-turbo        & 0.43 & 0.60 & 0.48 \\
GPT-4                & 0.53 & \textbf{0.69} & 0.58 \\
LLaMA-3-8B-Instruct  & 0.46 & 0.59 & 0.50 \\
LLaMA-3-70B-Instruct(Ours) & \textbf{0.60} & 0.68 & \textbf{0.62} \\
\bottomrule
\end{tabular}
\vspace{-1em}
\end{table}

\vspace{-1em}
\begin{table}[htbp]
\centering
\small
\renewcommand{\arraystretch}{2}
\caption{\textbf{Entity-Extraction Agent evaluation results.} Precision (Pr), Recall (Re), and F1 score under different similarity thresholds. Best results are bolded.}
\label{tab:2}
\resizebox{\columnwidth}{!}{%
\begin{tabular}{cccccccccc}
\hline
\multirow{2}{*}{\textbf{Method}} & \multicolumn{3}{c}{\textbf{Threshold = 60}}         & \multicolumn{3}{c}{\textbf{Threshold = 70}}         & \multicolumn{3}{c}{\textbf{Threshold = 80}}         \\ \cline{2-10} 
 & \textbf{Pr} & \textbf{Re} & \textbf{F1} & \textbf{Pr} & \textbf{Re} & \textbf{F1} & \textbf{Pr} & \textbf{Re} & \textbf{F1} \\ \hline
Stanza                           & 0.1689      & \textbf{0.3904}      & 0.2358      & 0.1373      & 0.3173      & 0.1917      & 0.1138      & 0.2630      & 0.1589      \\
BioEN                            & 0.2082      & 0.1065      & 0.1409      & 0.1469      & 0.0752      & 0.0994      & 0.1102      & 0.0564      & 0.0746      \\
LLM(Ours)          & \textbf{0.2120}      & 0.3528      & \textbf{0.2649}      & \textbf{0.1932}      & \textbf{0.3215}      & \textbf{0.2414}      & \textbf{0.1606}      & \textbf{0.2672}      & \textbf{0.2006}      \\ \hline
\end{tabular}
}
\end{table}

Table~\ref{tab:1} reports the results of the De-identification Agent evaluated against SOTA zero-shot baselines. While the rule-based PhysioNet baseline exhibits limited ability to identify sensitive entities, all LLM-based methods achieve significantly better performance. Among them, our design based on LLaMA-3-70B-Instruct outperforms all other models, supporting its use as the De-identification Agent.

Table~\ref{tab:2} shows the evaluation of the Entity-extraction Agent compared to SOTA baseline methods. Our model consistently outperforms both Stanza and BioEN in terms of F1 score across all similarity thresholds.  While Stanza achieves the highest recall, especially at lower thresholds, its precision remains lower than that of the Entity-extraction Agent, reflecting a higher number of false positives. BioEN performs well in precision at the lowest threshold but suffers from very low recall, likely due to its broader but less disease-focused entity coverage. As the fuzzy matching threshold increases, precision decreases for all methods, reflecting stricter matching with ontology terms. The Entity-extraction Agent maintains a strong balance between precision and recall, leading to the highest F1 scores across all thresholds.

\subsection{Hierarchical Ontology Extension: LLM Evaluation}
\label{sec:result2}

We evaluate the effectiveness of our Ontology Extension design against the baseline methods described in Section~\ref{sec:baselines}, using the evaluation metrics detailed in Section~\ref{sec:metrics}.

\vspace{-1em}
\begin{table}[htbp]
\centering
\small
\renewcommand{\arraystretch}{1.2}
\caption{\textbf{Hierarchical Ontology Extension LLM Evaluation Results.} Comparison of precision and the number of Correct, Incorrect, and Not Sure predictions across our method and three baselines. Best results are bolded.}
\label{tab:3}
\resizebox{\columnwidth}{!}{%
\begin{tabular}{ccccc}
\toprule
\textbf{Models} & \textbf{Correct} & \textbf{Incorrect} & \textbf{Not Sure} & \textbf{Precision} \\
\midrule
LLM-Onetime              & 293 & 461 & 99  & 38.859 \\
LLM-Hierarchical         & 268 & 463 & 53  & 36.662 \\
SapBERT+LLM-Onetime      & 241 & 318 & 322 & 43.113 \\
SapBERT+LLM-Hierarchical(Ours) & 617 & 158 & 106 & \textbf{79.613} \\
\bottomrule
\end{tabular}
}

\end{table}

Table~\ref{tab:3} presents the performance of our method's design (SapBERT+LLM-Hierarchical) compared to the three baselines. Our method achieves the highest precision, approaching 80\%, and produces the largest number of correct entity-node triplets. LLM-Onetime and LLM-Hierarchical suffer from incorrect placements, likely due to their lack of domain-specific embeddings and inability to reason over hierarchical structures. Although they produce fewer ambiguous cases labeled ``Not Sure'', they tend to misclassify the correct location, especially in deeper hierarchies. SapBERT+LLM-Onetime improves similarity matching via domain-specific embeddings but still lacks layer-wise reasoning, resulting in fewer correct placements compared to our proposed hierarchical ontology extension method.

These results highlight the importance of combining domain-specific semantic embedding (via SapBERT) and hierarchical reasoning (via LLM) to achieve accurate and scalable ontology extension.

\fussy

\subsection{Hierarchical Ontology Extension: Human Evaluation}
\label{sec:result3}
In addition to automated evaluations in Section \ref{sec:result2},  we conducted a comprehensive human evaluation to assess the semantic relevance and hierarchical accuracy of the extended ontology. Two annotators with expertise in biomedicine terminology, recruited from a nursing school affiliated with a major research university, participated in the evaluation process. From the ontology extension results produced by all four methods, we randomly sampled 100 new nodes that appeared in the outputs. For each sampled node, we extracted its hierarchical context and constructed evaluation triplets in the form: \textit{\{new\_node, node\_from\_original\_ontology, relationship\}}.

% \noindent Each triplet was rated by both annotators on a 3-point scale:
% \begin{itemize}
% \item 0 – the criterion was not met,
% \item 1 – partial alignment,
% \item 2 – full alignment.
% \end{itemize}

% \noindent Ratings were assigned based on four evaluation criteria:
% \begin{itemize}
% \item Relevance: whether the new entity is semantically appropriate within the disease ontology;
% \item Accuracy: whether the hierarchical relationship is correctly specified;
% \item Importance: whether the new entity is a meaningful addition to the ontology;
% \item Overall Satisfaction: overall quality and consistency of the triplet.
% \end{itemize}

\noindent Each triplet was rated by both annotators on a 3-point scale (0–2), where 0 indicates that the criterion was not met, 1 denotes partial alignment, and 2 represents full alignment. Ratings were assigned based on four evaluation criteria: \textit{Relevance} (whether the new entity is semantically appropriate within the disease ontology), \textit{Accuracy} (whether the hierarchical relationship is correctly specified), \textit{Importance} (whether the new entity is a meaningful addition to the ontology), and \textit{Overall Satisfaction} (the overall quality and consistency of the triplet).

\vspace{-1.5em}
\begin{table}[htbp]
\centering
\small
\renewcommand{\arraystretch}{1.2}
\caption{\textbf{Hierarchical Ontology Extension Human Evaluation Results.} Comparison across relevance (Rel.), accuracy (Acc.), importance (Imp.), and overall. Best results are bolded.}
\label{tab:4}
\resizebox{\columnwidth}{!}{%
\begin{tabular}{ccccc}
\toprule
\textbf{Models} & \textbf{Rel.} & \textbf{Acc.} & \textbf{Imp.} & \textbf{Overall} \\
\midrule
LLM-Onetime              & 0.09 & 0.67 & 0.63 & 0.06 \\
LLM-Hierarchical         & 0.09 & 0.96 & 0.63 & 0.09 \\
SapBERT+LLM-Onetime      & 1.04 & 1.53 & 1.00 & 1.00 \\
SapBERT+LLM-Hierarchical(Ours) & \textbf{1.05} & \textbf{2.00} & \textbf{1.87} & \textbf{1.87} \\
\bottomrule
\end{tabular}
}
% \vspace{-1.5em}
\end{table}

The results, summarized in Table~\ref{tab:4}, show that our proposed hierarchical ontology extension framework achieved the highest average ratings across all four evaluation metrics. Specifically, it outperformed the second-best model, SapBERT+LLM-Onetime, by an average of 51.2\% across the metrics. Both LLM-Onetime and LLM-Hierarchical methods showed notably low performance on the Relevance metric, highlighting a key limitation of using LLMs without domain-specific pretraining. Without adequate grounding in biomedical knowledge, these models struggle to capture semantically appropriate relationships, regardless of whether they incorporate hierarchy. Furthermore, our framework achieved the highest score on the Accuracy metric, demonstrating its ability to identify precise hierarchical relations between medical entities. 
\section{Conclusion}

This study presents \modelname, a novel zero-shot hierarchical ontology extension framework tailored for clinical notes, leveraging the power of LLMs to address key challenges including privacy-preserving preprocessing, hierarchical complexity, and data scarcity. 
% By leveraging a dual-agent system for entity extraction and de-identification, and incorporating SapBERT and Llama for semantic and structural consistency, 
Experimental results demonstrate that \modelname\ consistently outperforms baseline methods across multiple evaluation settings, highlighting its effectiveness in aligning new medical concepts with existing ontological structures.
Despite the promising results, the current evaluation is limited by a small dataset of 100 clinical notes due to the labor-intensive nature of privacy-compliant annotation. Ongoing efforts focus on expanding the dataset and improving methodological robustness. Future directions include enhancing scalability, interpretability, and alignment with clinical decision-making to support real-world applications in healthcare.

\bibliographystyle{IEEEtran}
\bibliography{references}

% \begin{thebibliography}{00}
% \bibitem{b1} G. Eason, B. Noble, and I. N. Sneddon, ``On certain integrals of Lipschitz-Hankel type involving products of Bessel functions,'' Phil. Trans. Roy. Soc. London, vol. A247, pp. 529--551, April 1955.
% \bibitem{b2} J. Clerk Maxwell, A Treatise on Electricity and Magnetism, 3rd ed., vol. 2. Oxford: Clarendon, 1892, pp.68--73.
% \bibitem{b3} I. S. Jacobs and C. P. Bean, ``Fine particles, thin films and exchange anisotropy,'' in Magnetism, vol. III, G. T. Rado and H. Suhl, Eds. New York: Academic, 1963, pp. 271--350.
% \bibitem{b4} K. Elissa, ``Title of paper if known,'' unpublished.
% \bibitem{b5} R. Nicole, ``Title of paper with only first word capitalized,'' J. Name Stand. Abbrev., in press.
% \bibitem{b6} Y. Yorozu, M. Hirano, K. Oka, and Y. Tagawa, ``Electron spectroscopy studies on magneto-optical media and plastic substrate interface,'' IEEE Transl. J. Magn. Japan, vol. 2, pp. 740--741, August 1987 [Digests 9th Annual Conf. Magnetics Japan, p. 301, 1982].
% \bibitem{b7} M. Young, The Technical Writer's Handbook. Mill Valley, CA: University Science, 1989.
% \end{thebibliography}
% \vspace{12pt}
% \color{red}
% IEEE conference templates contain guidance text for composing and formatting conference papers. Please ensure that all template text is removed from your conference paper prior to submission to the conference. Failure to remove the template text from your paper may result in your paper not being published.

\end{document}